\documentclass[final]{cvpr}

\usepackage{times}
\usepackage{epsfig}
\usepackage{graphicx}
\usepackage{amsmath}
\usepackage{amssymb}

\usepackage{subfigure}
\usepackage{subfiles}

\usepackage{algpseudocode}
\usepackage[ruled,vlined]{algorithm2e}
\usepackage{subfigure}
\usepackage{soul}
\usepackage{multirow}
\usepackage{xcolor,colortbl}
\definecolor{Gray}{gray}{0.9}
\newcolumntype{g}{>{\columncolor{Gray}}c}
\DeclareMathOperator*{\argmax}{argmax}

\usepackage[pagebackref=true,breaklinks=true,colorlinks,bookmarks=false]{hyperref}

\def\cvprPaperID{7389} 
\def\confYear{CVPR 2021}

\begin{document}

\title{Gallery Sampling for Robust and Fast \\Face Identification}

\author{Myung-cheol Roh, Pyoung-gang Lim, Jongju Shin\\
kakaoenterprise corp.\\
Seongnam, Kyonggi-do, South Korea
251-500\\
{\tt\small myung.roh@gmail.com, muzzynine@gmail.com, isaac.giant@kakaoenterprise.com}
}

\maketitle
\maketitle

\begin{abstract}
Deep learning methods have been achieved brilliant results in face recognition.
One of the important tasks to improve the performance is to collect and label images as many as possible.
However, labeling identities and checking qualities of large image data are difficult task and mistakes cannot be avoided in processing large data.
Previous works have been trying to deal with the problem only in \emph{training domain}, however it can cause much serious problem if the mistakes are in \emph{gallery data} of face identification.

We proposed gallery data sampling methods which are robust to outliers including wrong labeled, low quality, and less-informative images and reduce searching time.
The proposed sampling-by-pruning and sampling-by-generating methods significantly improved face identification performance on our $5.4$M web image dataset of celebrities.
The proposed method achieved 0.0975 in terms of FNIR at FPIR=0.01, while conventional method showed 0.3891. 
The average number of feature vectors for each individual gallery was reduced to 17.1 from 115.9 and it can provide much faster search.
We also made experiments on public datasets and our method achieved 0.1314 and 0.0668 FNIRs at FPIR=0.01 on the CASIA-WebFace and MS1MV2, while the convectional method did 0.5446, and 0.1327, respectively.
\end{abstract}


\section{Introduction}
Face recognition has been considered as one of the most interesting topics in computer vision.
Many deep learning models for face recognition have been published and they produced good feature vectors to recognize identities\cite{Ranjan2019AFA,chang2020data,Kim_2020_CVPR,vggface2,arcface}.
These models require enormous amounts of data and the more face images we collect and use, the better the performance of the model we can expect.
Face image data from all over the world is huge and still growing rapidly, but it is impossible to collect every individual's face images and train each individual.
Therefore, in general, a feature extraction model is trained using enormous amount of the collected data and the model is used for recognizing seen and unseen face images of mates (collected IDs) and non-mates (not collected IDs). 
In order to train a model, images and corresponding labels which indicate identities are required and quality check whether if the collected images are suitable for training is necessary. 
Labeling and quality checking of the collected images is in human hands. 
As the number of image data increases, deterioration in labeling and image quality is inevitable.
There are several approaches on dealing with the quality issue \emph{in training data}\cite{rolnick2018deep,learningtolearn,Jiang2018MentorNetLD,Lee2018CleanNetTL}.
However, it can cause much serious problem if wrong labelled IDs, low qualities of images, and less informative images are included \emph{in gallery/enrollment set} because they are directly connected to recognition performance : Suppose a query face is recognized as an gallery face that is labelled as a wrong ID.
Our interest in this paper is in dealing with the quality \emph{in the gallery set} rather than in the training set, for a given deep learning model.

Our target domain is face identification. 
The face identification is to find the most probable ID from a gallery set by comparing feature vectors and to decide whether to accept or reject the ID found.
For an instance, in celebrity identification, an user inputs a face image and wants to know who he/she is.
Numerous feature vectors of celebrities are enrolled in the celebrity database and the one whose distance is smallest is picked up with corresponding identity label.
However, as mentioned earlier, there would be many mislabelled IDs and inappropriate images in the dataset.
%
Under this condition, (a) developing a good feature extraction model is important and 
it is also very important (b) studying how to identify : how to retrieve the most probable ID in a huge feature space and how to decide whether to accept or reject the discovered identity.
Much of the previous papers' interest is in the former (a), but ours is in the latter (b).
In this paper, we present sampling methods for robust and fast identification.



The issues we want to address are as follows :
\begin{enumerate}
    \item The number of feature vectors in a gallery set : It is directly related to searching time as well as storage size.
    \item Qualities of gallery images and corresponding labels : The more image we collect, the larger variations of image quality and labeling quality. 
    \item Transitive property : 
    Let's say we have three different feature vectors, $a, b$ and $c$. The distance between two feature vectors, $a$ and $b$ are small enough to say they are same identity. 
    Also, the distance between $b$ and $c$ are small enough to say they are same identity.
    Does this always imply that $a$ and $c$ are the same identity?
\end{enumerate}

The first can be critical in real-world applications (e.g. world-wide celebrity's image searching).
The smaller number of gallery feature vectors, the faster searching time.

\begin{figure}[]
\centering
\begin{tabular}{@{}ccccccc@{}}
    \includegraphics[scale=0.20]{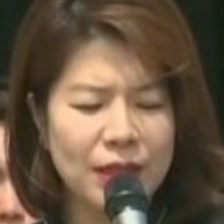}
    \includegraphics[scale=0.20]{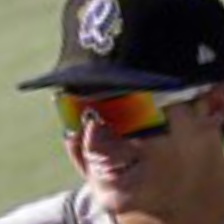}
    \includegraphics[scale=0.20]{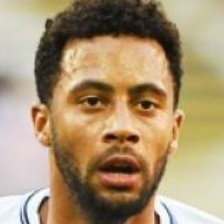}
    \includegraphics[scale=0.20]{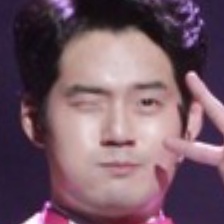}
    \includegraphics[scale=0.20]{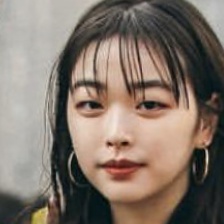}\\
    \includegraphics[scale=0.20]{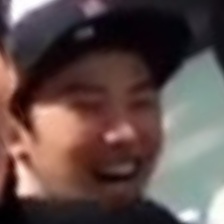}
    \includegraphics[scale=0.20]{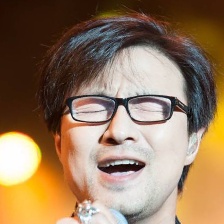}
    \includegraphics[scale=0.20]{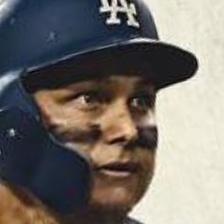}
    \includegraphics[scale=0.20]{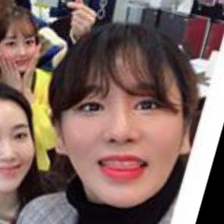}
    \includegraphics[scale=0.20]{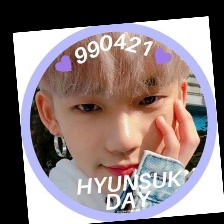}\\
    \includegraphics[scale=0.20]{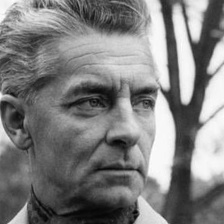}
    \includegraphics[scale=0.20]{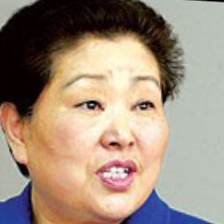}
    \includegraphics[scale=0.20]{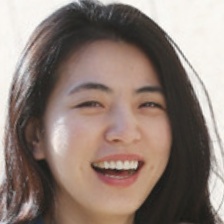}
    \includegraphics[scale=0.20]{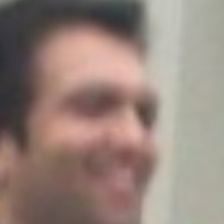}
    \includegraphics[scale=0.20]{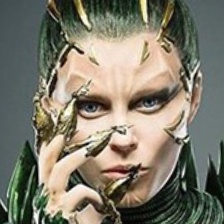}
\end{tabular}
\caption{Normalized face images crawled on the web have various conditions. Some of the images are not good to as reference images (i.e. gallery images) to identify}
\label{fig:dbimage}
\end{figure}

In order to enroll an image in a galley set, each image has to be labelled with its identity manually.
This is a tedious and difficult task to perform accurately.
Since the number of images is huge and many are likely to be unfamiliar faces, it happens to label incorrectly and incorrect label are found even after some iterations of verification processes.
Fig~\ref{fig:dbimage} shows our target face images crawled on the web.
Since face images on the web have limitless conditions, some images have bad qualities to recognize the identities.
Some faces are very hard to recognize its identity only by visual information without knowing additional contextual information because image resolution is very low or some facial parts are occluded.
These images are labelled using contextual information and they degrade the performance of face identification.
Therefore, maintaining the quality is important but it is difficult to set a clear guideline.

%


The third is that many researchers do not consider carefully and rely only on absolute distance between a query and a nearest gallery feature vectors.
Given a threshold $th$, two IDs of image $a$ and image $b$ are considered as the same if $\textit{distance} (F(a), F(b)) < th$ where $F$ is a feature extraction function of a DNN.
Assume that $a^I$ and $b^I$ are in our gallery and have the same identity, $I$.
Given an unknown query image $q$, if $\textit{distance} (F(a^I), F(q)) < th$, the $q$ is recognized as $I$, no matter how large the distance between $q$ and $b$ is.
This property may extend decision boundary upto unnecessary areas.
For an example, if $b$ is (near) at boundary of the identity, any $q$ which is beyond the boundary do not belong to the identity.
However, previous methods do not consider this case. 

In this paper, we propose sampling methods that remove outliers and generate small number of gallery samples which can represent boundary of each identity.
In this paper, the outliers represent wrong labelled images, low-resolution images, and less informative images that one can hardly recognize its identity unless contextual information is given.
The proposed method greatly improved face identification performance on a large dataset.
On a large dataset of 5.4M images, the proposed method achieved \emph{false negative identification rate (FNIR)}=$0.0975$ at \emph{false positive identification rate (FPIR)=0.01} while the conventional method achieved $0.3891$ FNIR at the same FPIR.
The average number of feature vectors of each IDs is reduced from $115.9$ to $17.7$.
Also, the proposed method achieved $0.1314$ and $0.0668$ FNIRs at FPIR=0.01 on the CASIA-WebFace and MS1MV2, respectively, while the convectional method did $0.5446$, and $0.1327$, respectively.

\section{Sampling}
Our method is to find a small set of feature vectors such that it can represent more efficiently its distribution robust to outliers in face identification.
Given a feature vector of a query face image, general face identification process is performed by calculating distances between the input feature vector (query) and enrolled feature vectors (gallery), finding a closest one, and making decision if it is suitable to assign the query as the identity of the closest one.

\begin{figure*}
\centering
\subfigure[Conventional distance measuring]{
    \includegraphics[scale=0.55]{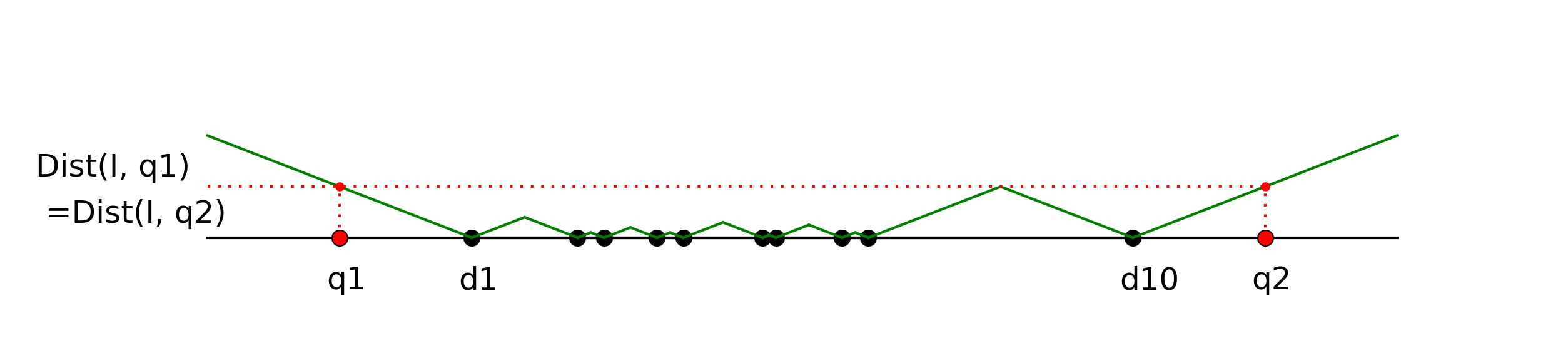}
    \label{fig:method:distancemeasurementA}}
\subfigure[Mahalanobis distance measuring]{
    \includegraphics[scale=0.55]{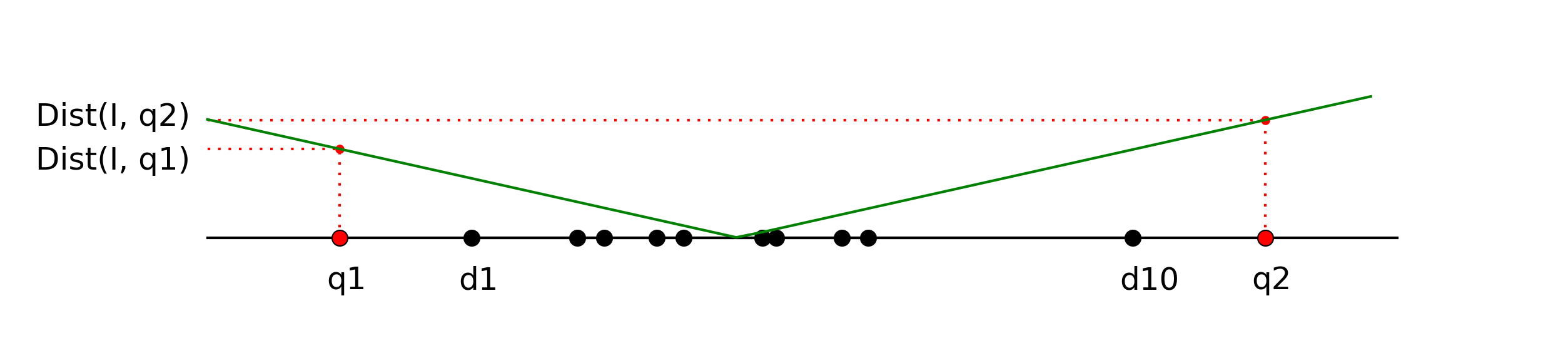}
    \label{fig:method:distancemeasurementB}}
\subfigure[Distance measuring using our sampling-by-generating]{
    \includegraphics[scale=0.55]{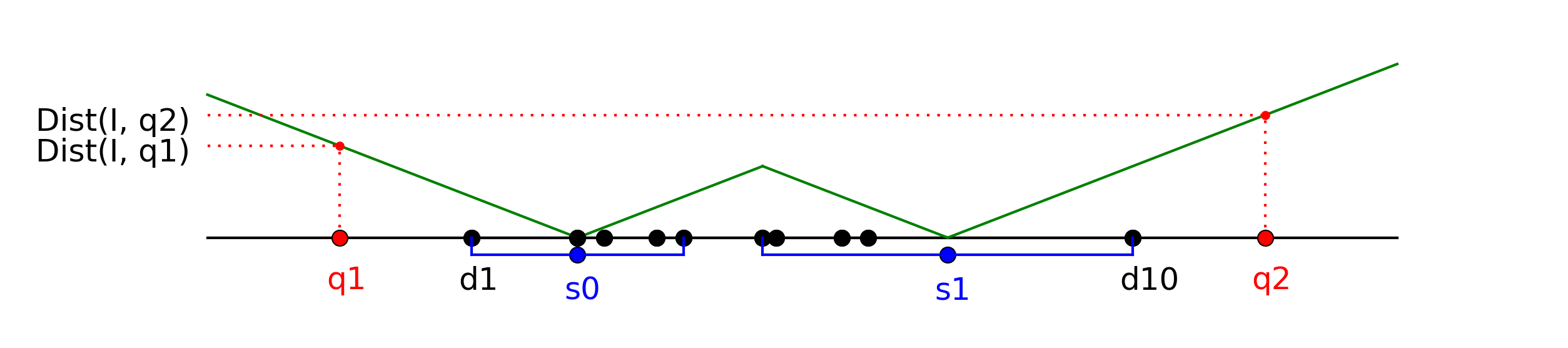}
    \label{fig:method:distancemeasurementC}}
\subfigure[Distance measuring using our sampling-by-pruning and sampling-by-generating]{
    \includegraphics[scale=0.55]{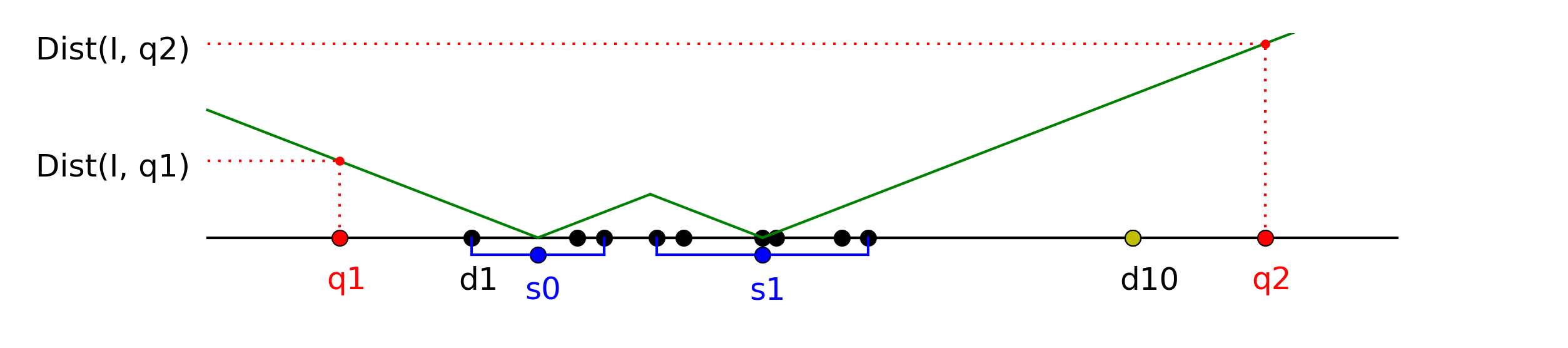}
    \label{fig:method:distancemeasurementD}}
\caption{Distance measuring. x and y axes represent 1D feature space and distance from \emph{a given identity}, respectively. 
Black and red dots represent enrolled and query(test) feature vectors, respectively. 
Blue dots represent new samples which is to be used as gallery data instead of $d_1 \cdots d_{10}$.
Green lines represent distance measuring functions.}
\label{fig:method:distancemeasurement}
\end{figure*}

Fig.~\ref{fig:method:distancemeasurement} shows distance measuring functions of conventional, Mahalanobis, and our methods.
$x$-axis represents feature space and $y$-axis represents distance from a gallery set. 
$d_1, \cdots, d_{10}$ are 1D gallery feature vectors which belong to an identity ($I$) and $q_1$ and $q_2$ are feature vectors of two query images.

Fig.~\ref{fig:method:distancemeasurementA} shows conventional distance function.
Conventionally the distance $Dist_{Cnv}(I, q)$ is defined by $min(dist(q, d_k))$ for $k = \{1, \cdots , 10\}$ where $dist$ is L2-norm.
The green lines represent Euclidean distance function, $Dist_{Cnv}(\cdot, \cdot)$.
The distance of $q_1$ from the identity $I$ is an Euclidean distance between $q_1$ and $d_1$, which is $Dist_{Cnv}(I, q_1)$.
Let a threshold $th$ is given where $dist(q_i, q_j) < th$ represent that $q_i$ and $q_j$ have the same identity.
Under the conventional measurement, for a given threshold $th$, if
\begin{align}
    dist(d_0, d_1) &< th, \nonumber\\
        \vdots&\nonumber  \\
    dist(d_9, d_{10}) &< th,  \nonumber
\end{align}
and 
\begin{align}
    dist(d_{10}, q_{2}) &< th, \nonumber
\end{align}
is it trivial to say that $d_1$ and $q_0$ have the same identity?
Since $d_{10}$ is an isolated data from others, it is likely to be an outlier or a boundary data of the identity.
It could be come from low-resolution image or wrong-labeled image.
The distribution of feature vectors is not considered in the conventional distance measurement and it cannot deal with the outlier.

In \cite{NIPS2018_7947}, they proved that the softmax based classifier is equivalent to fitting the class-conditional Gaussian distributions.
Therefore, a set of trained feature vectors of an identity, $I$ is normally distributed, $\mathcal{N}(\mu_i, \Sigma_i)$, and if a model is trained using large enough data we can expect the theorem will work on unseen data as well.
Based on the theorem, the distance can be calculated using Mahalanobis distance.
Given $q$ and $d_1, \cdots, d_{10}$, the distance measurement based on Mahalanobis is defined as follows:
\begin{align}
    Dist_{Mhl}(I, q) & = \sqrt{(q-\mu_I)^T{\Sigma_I}^{-1}(q-\mu_I)} \nonumber
\end{align}
where $\mu$ and $\Sigma$ are a mean vector and a covariance matrix of $d_1, \cdots, d_{10}$.
Fig.~\ref{fig:method:distancemeasurementB} shows the distance measurement using the Mahalanobis distance.
This measurement shows that $Dist_M(I, q_2)$ is larger than $Dist_M(I, q_1)$ as we expected.
If estimating the best parameters of function $\mathcal{N}(\mu_i, \Sigma_i)$ is always guaranteed it is good choice to use Mahalanobis distance.
However, because estimating the $\mu_i$ and $\Sigma_i$ requires large number of images for each identity and all the identities does not have large enough, it is not generally applicable.
Moreover, the mean and the variance of the collected data are sample mean and sample variance, thus some outliers can greatly influence the mean and variance estimates.



Our method generates new samples which can represent each identity's distribution/boundary and also it prunes isolated data which is assumed to be outliers.
Generating some representative samples and pruning outliers introduce a good distance measurement.
Fig.~\ref{fig:method:distancemeasurementC} shows the distance function of the proposed sampling-by-generating.
Two new feature vectors, $s_0$ and $s_1$, are generated from ($d_1, \cdots , d_5$) and ($d_6, \cdots , d_{10}$), respectively.
Since the distance is calculated using the carefully generated samples instead of all the data, it prevents the defect which comes from the transitive property we mentioned before.
As a result, it seems to use two Mahalanobis functions, but it does not need as many feature vectors as the Mahalanobis needs for estimating parameters.
Fig.~\ref{fig:method:distancemeasurementD} shows the distance function of the proposed sampling-by-pruning and -generating together.
In this method, two new feature vectors, $s_0$ and $s_1$, are generated from ($d_1, \cdots , d_3$) and ($d_4, \cdots , d_{9}$), respectively.
$d_{10}$ (a yellow dot) which is isolated is pruned and does not contribute to the generation.


\subsection{Sampling-by-generating}
The idea of the sample generation is for each identity to find centers of hyperspheres of radius $r$ such that the hyperspheres cover the feature vectors area.
The variance of the distribution is reflected by using several hyperspheres.
Thus, it can represent the features distribution by examplars and it can also represent complex distribution which is not following the Gaussian distribution.
The parameter $r$ is expected to set to 
\begin{align}
    \max (\{r \mid \forall v_a, v_b, \text{dist}(v_a, v_b) < r \implies  \nonumber \\
    v_a \text{ and } v_b \text{ are the same identity}\}).\nonumber
\end{align}
However, the $r$ will be too small under noisy condition, so we found the value empirically.
See the Ablation study for the analysis.
In the Fig.~\ref{fig:method:distancemeasurementC}, the blue dots are the generated samples which are mean vectors of two clusters : $\{d_1, \cdots , d_5\}$ and $\{d_6, \cdots , d_{10}\}$.
 For each cluster, the distance between a data and one of the the generated samples is $r$ at most: For any  $d_i \in \{d_1, \cdots , d_{10}\}$, 
\begin{align}
    \min_j \text{dist}(d_i, s_j) \leq r,\text{ where } s_j \in \{s_1, s_2\}.\nonumber
\end{align}

Detailed processes are shown in algorithm~\ref{algorithm:samplegeneration}.
This process is performed for each identity.
%
\begin{algorithm}[]
\SetAlgoLined
\KwResult{Sample set, $S$}
 Let $D$ is a set of $n$ feature vectors in a gallery set.\;
 Given hyper-sphere's radius $r$ and margin $\epsilon$,\;
 $S = \emptyset$\;
 $v_{init} = \text{mean}(v_1, \cdots , v_n)$, \;
 $v_1$ = $\argmax_{v_j \in D} \text{dist}(v_{init}, v_j)$ \;
 \While{$D$ is not empty}{
  $I_0$ = $\{ v_j | \text{dist}(v_1, v_j) < 2r - \epsilon, v_j \in D \}$\;
  $v_2 = \argmax_{v_k \in I_0} \text{dist}(v_1, v_k)$ \;
  $v_m = \text{mean}(v_1, v_2)$  \Comment{A generated sample}\;
  $I_1$ = $\{ v_j | \text{dist}(v_m, v_j) < r, v_j \in D\}$\Comment{Inliers}\;
  Move the $v_m$ toward $v_{init}$ while $I_1 \subset \{ v_j | \text{dist}(v_m, v_j) < r, v_j \in D\}$\; 
  $S = S \cup \{v_m\}$\; 
  $I_2$ = $\{ v_j | \text{dist}(v_m, v_j) < r, v_j \in D\}$\Comment{Inliers}\;
  $D = D - I_2$\;
  $v_1$ = $\argmax_{v_j \in D} \text{dist}(v_m, v_j)$ \;
  
 }
 \caption{Algorithm to generate samples}
 \label{algorithm:samplegeneration}
\end{algorithm}



\subsection{Sampling-by-pruning}
Pruning is to remove outliers and it is applied before the sampling-by-generating.
The feature vectors of outliers are tend to be far from good feature vectors and scattered meanwhile good feature vectors are dense.
Under an assumption that inliers are the majority, the most of the outliers can be pruned by removing small clusters.
The pruning is done by removing small clusters and clustering can be done by mean-shift clustering with pruning ratio and bandwidth as parameters.
The pruning ratio, $pr$, determine how many data will be pruned and the bandwidth, $b$ is the size scale of a kernel function in the mean-shift clustering.
In Fig.~\ref{fig:method:distancemeasurementD}, $d_{10}$ is pruned and two new feature vectors, $s_0$ and $s_1$ using the sampling-by-generating without using $d_{10}$.
%
%



\section{Experiments}
We collected about $5.4$M web-images for experiments.
$4,720,831$ images of $41,109$ identities are used for gallery set.
$626,955$ images of mates whose identities are in the gallery set and $54,730$ images of non-mates whose identities are not in the gallery set are used for probe (query) data.
We collected maximum 30 images per person for the probe set.
Fig.~\ref{fig:dbimage} shows some images we collected and it shows that the conditions of the images taken are various.
We labelled the images by hands through multiple screenings. 
Although we had multiple screenings, mislabelled images and low quality images still remain.

Also, we measured our methods on public datasets: CASIA-WebFace\cite{casia} and MS1MV2\cite{ms1mv2}.
For each dataset, we created 5 random splits of the dataset into gallery, mate probe, and non-mate probe sets.
For each dataset, $80\%$ of identities are randomly chosen as a mate set and $20\%$ are chosen as a non-mate probe set.
Among the mate sets, $80\%$ of images are randomly chosen as a gallery set and the others are chosen as a mate probe set.

We have conducted experiments using different feature sets: using raw (conventional) features (\textbf{Raw}), raw features after pruning (\textbf{PrunRaw}), generated sample features without pruning (\textbf{Gen}), and generated sample features after pruning (\textbf{PrunGen}).
In addition, we also tested a single feature computed by aggregating (averaging) the face feature vector of each identity\cite{vggface2} with and without pruning (\textbf{PrunSgl} and \textbf{Sgl}, respectively).

The feature vectors were obtained through the pretrained ArcFace model\cite{arcface} with Resnet100 which are publicly available in \cite{arcface_github}.

\subsection{Performance metric}
The performance is measured using False negative identification rates (FNIRs) at given False positive identification rates (FPIRs) as follows\cite{FRVT}: 

\begin{equation}
    \textbf{FNIR} =  \frac{\text{Num. Negatives}_{\textit{Mates}}}{\text{Num. Mates}}
\end{equation}
\begin{equation}
    \textbf{FPIR} =  \frac{\text{Num. Positives}_{\textit{Non-mates}}}{\text{Num. Non-mates}}
\end{equation}
Face identification can be applied to automatic face image tagging/classification and our target is celebrity face identification.
For a given query face image, if the subject is enrolled in the gallery set, the subject's ID/name is returned otherwise, {\it unidentified-subject} message is returned.
Thus, we considered top-1 accuracy.

In addition, precision and recall rates are measured and they are as follows:
\begin{equation}
    \textbf{Precision} =  \frac{\text{TP}}{\text{TP} + \text{FP}}
\end{equation}
\begin{equation}
    \textbf{Recall} =  \frac{\text{TP}}{\text{TP} + \text{FN}}
\end{equation}
where TP, FP and FN stand for True Positive, False Positive and False Negative.
To count the TP, FP, and FN, all mate and non-mate data are used.

\subsection{Results}
In our experiments, the parameters (radius $r$, bandwidth $b$, and pruning ratio $pr$) were found at where the FNIR is minimized at FPIR=$0.01$.
The FPIR was measured on non-mates data.
Table~\ref{tab:result:fnir} shows that the FNIR significantly decreased using either sampling-by-pruning or sampling-by-generation.
At FPIR$=0.01$, the FNIR decreased from $0.389$ (\textbf{Raw}) to $0.0975$ (\textbf{PrunGen}).
It also shows that the conventional \textbf{Raw} and \textbf{Sgl} methods are improved using the proposed sampling-by-pruning.

\begin{table}
\centering 
\caption{FNIRs according to different methods and FPIRs. The lower the better.}
 \begin{tabular}{c |c c g c c} 
 \hline
 \multirow{2}{*}{Methods} & \multicolumn{5}{c}{FPIRs} \\
  & $0.001$ & $0.005$ & $0.01$ &  $0.05$  &  $0.1$\\ 
 \hline \hline
 Raw      &   0.9064   &   0.5956   &   0.3891  &   0.0676  &  0.0408 \\
 PrunRaw  &   0.8959   &   \bf{0.3277}   &   0.1231  &   0.0700  &  0.0569 \\  
 Sgl      &   0.8650   &   0.4781   &   0.2630  &   0.2153  &  0.2046 \\  
 PrunSgl  &   0.8312   &   0.3963   &   0.1185  &   0.0645  &  0.0526 \\  
 Gen      &   \bf{0.7988}   &   0.3864   &   0.1231  &   \bf{0.0474}  &  \bf{0.0357} \\  
 PrunGen  &   0.8473   &   0.3948   &   \bf{0.0975}  &   0.0538  &  0.0441 \\ 
  \hline
\end{tabular}
 \label{tab:result:fnir}
\end{table}


Table~\ref{tab:result:prec} shows performance in terms of precision and recall measurement.
At the FPIR$=0.01$, precision and recall increased using the \textbf{PrunGen}.
Especially, recall was greatly improved from $0.6101$ to $0.9022$. 
In overall, using either sampling-by-generating or sampling-by-pruning improves the performance greatly and using both of them improve the performance even more.

\begin{table}
\centering 
 \caption{Precision and recall according to different methods at FPIR$=0.01$. The higher the better.}
\begin{tabular}{c |c c } 
 \hline
 Methods & Precision & Recall \\
 \hline\hline
 Raw     &   0.9951 & 0.6101\\
 PrunRaw  &  0.9956 & 0.8766\\
 Sgl     &   0.9955 & 0.7364\\
 PrunSgl  &   0.9958 & 0.8812\\
 Gen      &   0.9960 & 0.8766\\
 PrunGen  &   0.9959 & 0.9022\\

 \hline
\end{tabular}
\label{tab:result:prec}
\end{table}


Search time is proportional to the size of the gallery set.
Table~\ref{tab:result:numsamples} shows that the average numbers of gallery feature vectors of each identity.
\textbf{Sgl} aggregates by averaging feature vectors, thus the number of feature vectors of each identity is 1.
The \textbf{PrunGen} method reduces the number by 0.15 times and increases the accuracy with a large margin compared to the conventional method (\textbf{Raw}).

\begin{table}
\centering 
\caption{The average number of gallery feature vectors for each identity.}
 \begin{tabular}{c | c} 
 \hline
    method & number of gallery features\\
 \hline\hline
 Raw & 115.9 \\
 PrunRaw & 99.0\\
 Sgl & 1.0\\
 PrunSgl & 1.0\\ 
 Gen & 21.6\\
 PrunGen & 17.7 \\ 
 \hline
\end{tabular}
\label{tab:result:numsamples}
\end{table}

Tables~\ref{tab:result:fnir_public}, ~\ref{tab:result:prec_public}, ~\ref{tab:result:recall_public} show results on on the public datasets: MS1MV2 and CASIA-WebFace. 
For the experiments, we used same parameters as we used in our datasets.
From the tables, we can observe that the proposed either \textbf{Prun} or \textbf{Gen} methods outperformed others not only in the average, but also in the standard deviation.
It shows the proposed method works more reliably for data distributions and outliers.

\begin{table}
\centering 
\caption{FNIRs at FPIR=$0.01$ on CASIA and MS1MV2.}
 \begin{tabular}{c |c c | c c} 
 \hline
    \multirow{2}{*}{Methods} & \multicolumn{2}{c}{MS1MV2} &  \multicolumn{2}{c}{CASIA-WebFace} \\
           &   mean &  std    &  mean &    std \\
   \hline\hline
   Raw     & 0.1327 & 0.0355 & 0.5446 & 0.0234  \\
   PrunRaw & 0.0892 & 0.0561 & 0.1398 & 0.0029 \\
   Sgl     & 0.0970 & 0.0342 & 0.4381 & 0.1375 \\
   PrunSgl & 0.1059 & 0.0391 & 0.1363 & 0.0028 \\
   Gen     & 0.0673 & \bf{0.0228} & 0.2082 & 0.0129 \\
   PrunGen & \bf{0.0668} & 0.0256 & \bf{0.1314} & \bf{0.0025}  \\
  \hline
\end{tabular}
 \label{tab:result:fnir_public}
\end{table}

\begin{table}
\centering 
\caption{Precisions and recalls at FPIR=$0.01$ on CASIA-WebFace.}
 \begin{tabular}{c |c c | c c} 
 \hline
 \multirow{2}{*}{Methods} & \multicolumn{2}{c}{Precision} &  \multicolumn{2}{c}{Recall} \\
           &   mean &  std    &  mean &    std \\
   \hline\hline
    Raw     &  0.9892  & 0.0005  &  0.8671 & 0.0342  \\
    PrunRaw &  0.9899  & 0.0003  &  0.9106 & 0.0391  \\
    Sgl     &  0.9895  & 0.0001  &  0.9029 & 0.0355  \\
    PrunSgl &  0.9896  & 0.0003  &  0.8939 & 0.0561  \\
    Gen     &  0.9899  & 0.0002  &  0.9326 & 0.0228  \\
    PrunGen &  0.9899  & 0.0001  &  0.9330 & 0.0256  \\
  \hline
\end{tabular}
 \label{tab:result:prec_public}
\end{table}

\begin{table}
\centering 
\caption{Precisions and recalls at FPIR=$0.01$ on MS1MV2.}
 \begin{tabular}{c |c c | c c} 
 \hline
    \multirow{2}{*}{Methods} & \multicolumn{2}{c}{Precision} &  \multicolumn{2}{c}{Recall} \\
           &   mean &  std    &  mean &    std \\
   \hline\hline
   Raw      & 0.9625  & 0.0016   & 0.4540  &  0.0234  \\
   PrunRaw  & 0.9746  & 0.0004   & 0.8592  &  0.0029  \\
   Sgl      & 0.9592  & 0.0139   & 0.5589  &  0.1386  \\
   PrunSgl  & 0.9751  & 0.0003   & 0.8627  &  0.0027  \\
   Gen      & 0.9749  & 0.0004   & 0.7907  &  0.0129  \\
   PrunGen  & 0.9750  & 0.0003   & 0.8676  &  0.0025  \\
  \hline
\end{tabular}
 \label{tab:result:recall_public}
\end{table}




\section{Ablation Study}
We will discuss performance changes according to various parameter settings of our methods in order to understand how the parameters work.
The parameters are radius ($r$) in the sampling-by-generating and bandwidth ($bw$) and pruning ratio ($pr$) in the sampling-by-pruning.

\subsection{Sampling-by-pruning : bandwidth and pruning ratio}
\begin{figure}[h!]
\centering
\includegraphics[scale=0.5]{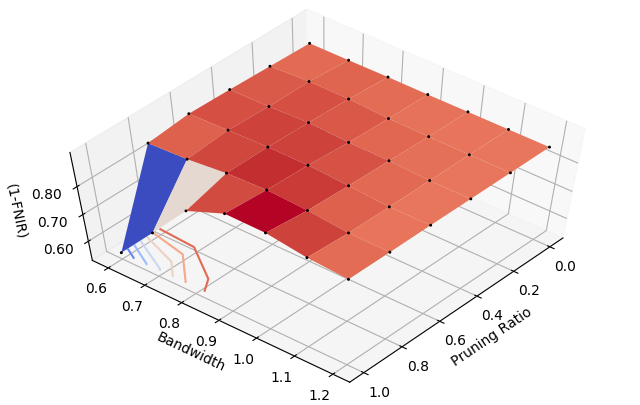}
\caption{FNIR according to bandwidth and pruning ratio}
\label{fig:ablation:pruning}
\end{figure}

In this experiment, we measure the performance according to the pruning ratio and the bandwidth where the radius is fixed.
The pruning ratio ($p$) varies from $0\%$ to $100\%$ and the bandwidth $b$ varies from from $0.6$ to $1.1$.
Note that the pruning ratio  $0.0\%$ represents that no pruning is applied and $100.0\%$ represents all clusters are pruned except the largest one.
Fig.~\ref{fig:ablation:pruning} shows the result.
The changes of the pruning ratio does not make significant changes in the performance except in where $b \leq 0.8$.
The results along the $pr = 1.0$ show better results than others except in where $b \leq 0.8$.

\begin{figure}[h]
\centering
\begin{tabular}{@{}ccccccc@{}}
    \includegraphics[scale=0.20]{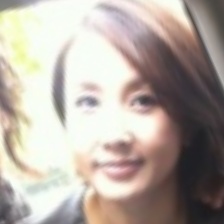}
    \includegraphics[scale=0.20]{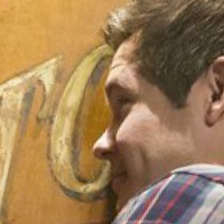}
    \includegraphics[scale=0.20]{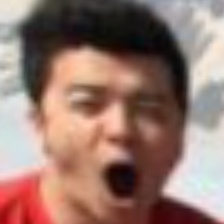}
    \includegraphics[scale=0.20]{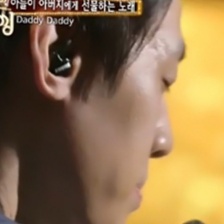} 
    \includegraphics[scale=0.20]{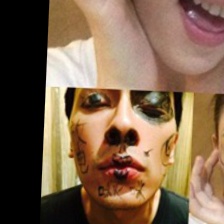} \\
    \includegraphics[scale=0.20]{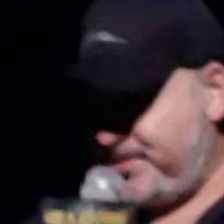}
    \includegraphics[scale=0.20]{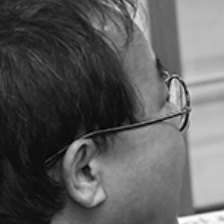}  
    \includegraphics[scale=0.20]{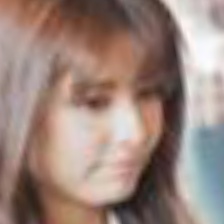}
    \includegraphics[scale=0.20]{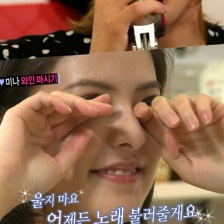}
    \includegraphics[scale=0.20]{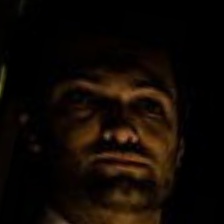}\\
    \includegraphics[scale=0.20]{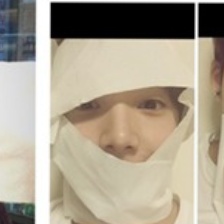}
    \includegraphics[scale=0.20]{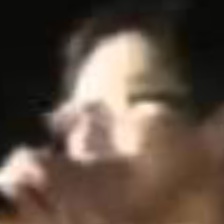}
    \includegraphics[scale=0.20]{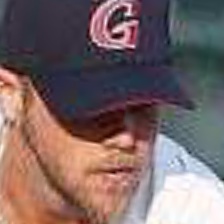}
    \includegraphics[scale=0.20]{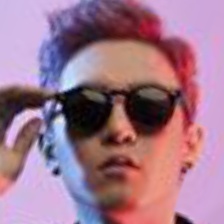}
    \includegraphics[scale=0.20]{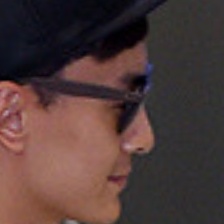}
\end{tabular}
\caption{These images are outlier images which are pruned by the proposed sampling-by-pruning.}
\label{fig:outliers}
\end{figure}

Fig.~\ref{fig:outliers} shows outlier images which are pruned from our galley set.
Because of occlusion and rotation of face and low resolution of image, these faces are difficult to recognize its identities.
The outliers are from the gallery set, and it also shows that outlier images could not be treated properly even after some iterations of quality verification.

\subsection{Sampling-by-generation}
\begin{figure}[h!]
\centering
\includegraphics[scale=0.5]{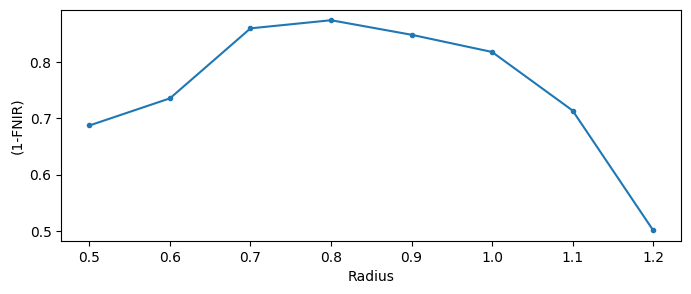}
\caption{(1-FNIR) according to the radius $r$.}
\label{fig:ablation:sampling}
\end{figure}

In the sampling-by-generation method, the only parameter is the radius ($r$).
Fig.~\ref{fig:ablation:sampling} shows the experimental results for $0.5 \leq r \leq 1.2$.
Its trend is similar to the graph in Fig.~\ref{fig:ablation:pruning} along $bw = 1.2$.
The radius is related to the number of samples generated. 
As the radius increases, the number of samples decreases. 
For an instance, average sample numbers are 36.0 and 1.0 at $r=0.5$ and $r=1.2$, respectively. 

\subsection{Radius and Bandwidth}
We measured the performance according to the radius $r$ and bandwidth $bw$ in the sample generation where pruning ratio is fixed.
Table~\ref{tab:ablation:radiusbandwidth} shows the parameters for the best top ten.
The lowest FNIR is achieved at where  $pr= 1.0$, $b = 0.9$, and $r = 0.7$.
Bandwidths $b$ of mean-shift clustering in top 4 are all $0.9$ and the majority value for radius $r$ is $0.7$.
For pruning ratio $pr$, 7 out of 10 are $1.0$.

\begin{table}
\centering 
\caption{PrunGen's top 10 parameter settings according to FNIR.}
 \begin{tabular}{c@{\hskip 0.1in} c@{\hskip 0.1in} c@{\hskip 0.1in} | c } 
 \hline
  $pr$ & $b$ & $r$ & FNIR\\
 \hline\hline
1.0 & 0.9 & 0.7 &    0.0975  \\
1.0 & 0.9 & 0.7 &    0.0978  \\
1.0 & 0.9 & 0.8 &    0.1030  \\
1.0 & 0.9 & 0.6 &    0.1033  \\
1.0 & 1.0 & 0.7 &    0.1074  \\
1.0 & 0.8 & 0.6 &    0.1129  \\
0.8 & 0.9 & 0.7 &    0.1134  \\
1.0 & 0.8 & 0.7 &    0.1135  \\
0.6 & 0.9 & 0.7 &    0.1158  \\  
0.8 & 0.8 & 0.7 &    0.1187  \\  
  \hline
\end{tabular}
\label{tab:ablation:radiusbandwidth}
\end{table}

\begin{figure}[h!]
\centering
\includegraphics[scale=0.55]{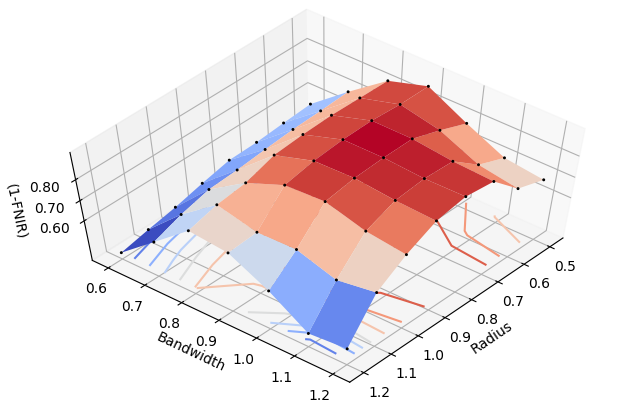}
\caption{(1-FNIR) according to bandwidth and radius parameters.}
\label{fig:ablation:radiuswidthFPIR}
\end{figure}

Fig.~\ref{fig:ablation:radiuswidthFPIR} shows $(1-FNIR)$ graph with respect to the bandwidth and radius given pruning ratio=1.0.
We can see that the accuracy falls smoothly off the peak and it does not show sharp change for small changes in parameter values.
Therefore, the best parameter could be found from some sampling parameter set or by some simple parameter estimation methods such as gradient descent method.


\subsection{Performance according to model depth }
\begin{figure}[h]
\centering
\subfigure[Resnet100]{
    \includegraphics[scale=0.70]{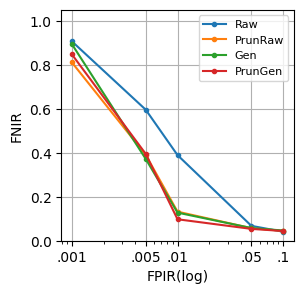}
    \label{fig:resnet100}}
\subfigure[Resnet50]{
    \includegraphics[scale=0.70]{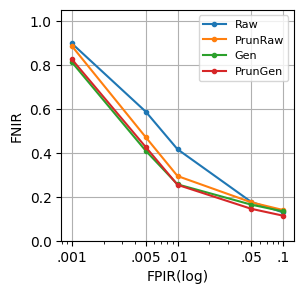}
    \label{fig:resnet50}}
\subfigure[Resnet34]{
    \includegraphics[scale=0.70]{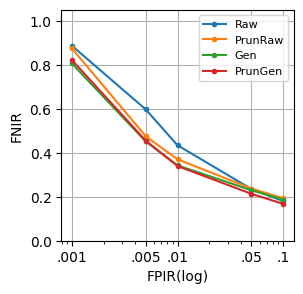}
    \label{fig:resnet34}}
\caption{FPIR-FNIR graphs according to different backbones of ArcFace models.}
\label{fig:ablation:resnet}
\end{figure}

In this experiment, the performance changes according to the model depth are measured.
The pretrained Resnet100, Resnet50, and Resnet34 models\cite{arcface_github} are used for this experiment.
Fig.~\ref{fig:ablation:resnet} shows FPIR-FNIR graphs according to the different models.
The proposed methods upon all the models outperform the conventional method.

An interest thing is the performances of \textbf{Raw}s for the different models do not show significant differences in between FPIR=0.001 and FPIR=0.01.
However, using either sampling-by-pruning or sampling-by-generating makes big differences.
The differences between \textbf{Raw} and \textbf{PrunGen} at FNIR=0.01 are $0.0943$ for Renet34,  $0.1616$ for Renet34, and $0.2916$ for Renet100.
It shows that although different models show seemingly similar results with conventional feature vectors as it is they have different potentials which may help to increase additional performance.

\section{Conclusion}
Labeling and checking qualities of images are difficult task and mistakes cannot be avoided in processing large data.
There are several approaches to deal with this problem in \emph{training domain}, however, it can cause much serious problem when the mistakes are in a \emph{gallery set}.
We proposed sampling-by-pruning and sampling-by-generating methods which are robust to outliers including wrong labeled, low quality, and less-informative images and reduce searching time by using small set of feature vectors as a gallery set. 
As a result, the sampling method greatly improved face identification performance in 5.4M dataset.
The proposed method achieved $0.0975$ while conventional feature vectors achieved $0.3891$ in terms of FNIR at FPIR=0.01. 
The average number of feature vectors for each individual gallery is reduced to 17.1 from 115.9 and it provides much faster search.
Also, the proposed method achieved $0.1314$ and $0.0668$ FNIRs at FPIR=0.01 on the CASIA-WebFace and MS1MV2, respectively, while the convectional method did $0.5446$, and $0.1327$, respectively.


{\small
\bibliographystyle{ieee_fullname}
\bibliography{references}
}

\end{document}